\documentclass{webofc}
\usepackage[varg]{txfonts}   
\makeatletter
\renewcommand{\@biblabel}[1]{#1.}
\makeatother

\DeclareUnicodeCharacter{202F}{(FIX ME!!!!)}

\begin{document}
\title{The Application of Artificial Neural Network Model to Predicting the Acid Mine Drainage from Long-Term Lab Scale Kinetic Test}
%
%

\author{\firstname{Muhammad Sonny} \lastname{Abfertiawan}\inst{1} \and
        \firstname{Muchammad Daniyal} \lastname{Kautsar}\inst{2} \and
        \firstname{Faiz} \lastname{Hasan}\inst{3} \and
        \firstname{Yoseph} \lastname{Palinggi}\inst{4} \and
        \firstname{Kris} \lastname{Pranoto}\inst{4}
}

\institute{Water and Wastewater Research Group, Faculty of Civil and Environmental Engineering, Bandung Institute of Technology 
\and Department of Electrical and Information Engineering, Universitas Gadjah Mada
\and Environmental Engineering Master Program, Faculty of Civil and Environmental Engineering, Bandung Institute of Technology
\and Environmental Department, PT Kaltim Prima Coal, Indonesia}

\abstract{%
  Acid mine drainage (AMD) is one of the common environmental problems in the coal mining industry that was formed by the oxidation of sulfide minerals in the overburden or waste rock. The prediction of acid generation through AMD is important to do in overburden management and planning the post-mining land use. One of the methods used to predict AMD is a lab-scale kinetic test to determine the rate of acid formation over time using representative samples in the field. However, this test requires a long-time procedure and large amount of chemical reagents lead to inefficient cost. On the other hand, there is potential for machine learning to learn the pattern behind the lab-scale kinetic test data. This study describes an approach to use artificial neural network (ANN) modeling to predict the result from lab-scale kinetic tests. Various ANN model is used based on 83 weeks experiments of lab-scale kinetic tests with 100\% potential acid-forming rock. The model approaches the monitoring of pH, ORP, conductivity, TDS, sulfate, and heavy metals (Fe and Mn). The overall Nash-Sutcliffe Efficiency (NSE) obtained in this study was 0.99 on training and validation data, indicating a strong correlation and accurate prediction compared to the actual lab-scale kinetic tests data. This show the ANN ability to learn patterns, trends, and seasonality from past data for accurate forecasting, thereby highlighting its significant contribution to solving AMD problems. This research is also expected to establish the foundation for a new approach to predict AMD, with time efficient, accurate, and cost-effectiveness in future applications.
  %
}
\maketitle
\section{Introduction}
For many years, Artificial Neural Networks (ANNs) have emerged as an incredible tool, celebrated for their remarkable capability and robustness when dealing with complex datasets. ANNs, inspired by the human brain's neural structure \cite{ref1-01}, have witnessed rapid and major developments, proving their capabilities across many domains and setting new benchmarks as State-of-the-Art (SOTA) models in various modalities.

The potential of ANNs lies in their ability to comprehend intricate patterns and relationships within data, often transcending the limits of conventional analytical techniques. This paper explores whether this ability of ANNs can be implemented in the environmental engineering domain. Specifically, to investigate their utility in understanding and forecasting Acid Mine Drainage (AMD) through Long-Term Lab Scale Kinetic Tests.

AMD is strong, acidic wastewater rich in high concentrations of dissolved ferrous and non-ferrous metal sulfates and salts, and if AMD is left untreated, it can contaminate ground and surface watercourses, damaging the health of plants, humans, wildlife, and aquatic species \cite{ref1-02}. Traditionally, AMD is analyzed using the kinetic test. This process involves simulating mine drainage production from samples influenced by mining activities, and incorporates dynamic elements encompassing physical, chemical, and biological systems and processes that govern the generation of acidic or alkaline mine drainage. Kinetic tests primarily focus on studying reaction rates and mechanisms leading to acidic or alkaline mine drainage, typically requiring larger sample volumes and extended durations than static tests, and are typically conducted in laboratory settings. These tests yield crucial information regarding sulfide mineral oxidation rates, acid production rates, and drainage water quality \cite{ref1-03}. Usually, comprehending the complex kinetics of AMD generation and forecasting its behavior through kinetic tests has been laborious and resource-intensive. Consequently, developing cost-effective and sustainable remediation solutions for the AMD problem has been the subject of extensive research. 

Predicting and forecasting analysis using ANN or Machine learning (ML) in general, has been used in various domain, including nonlinear timeseries and have gained overwhelming attention over the past years \cite{ref2-01}. Forecasting mining influenced water data using various ML technique including tree based method and ANN show positive result, with close to accurate prediction \cite{ref2-02}. Beside ML method, traditional model such as auto regressive integrated moving average (ARIMA), Box-Jenkins, etc. has been applied to this domain problem, but also coming with some drawback since traditional model assume that time series data are linear processes \cite{ref2-03}. This study will focus on developing ANN model with various technique that suits the time series domain and will focus on capturing the relation between data over time.

\section{Methodology}

\subsection{Feedforward Neural Network (FNN)}

\begin{figure}[ht]
\centering
\includegraphics[width=0.9\linewidth]{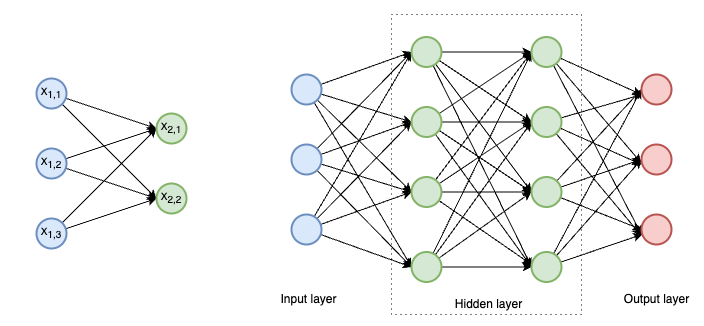}
\caption{(Left) Interconnected neurons. (Right) Feedforward network architecture.}
\label{fig3-1}
\vspace{-18pt} 
\end{figure}

One of the most common type of ANN architecture is a feedforward neural network (FNN). This type of model consists of several layers of linear model, with each corresponding number of neurons. A linear model (Eq. ~\ref{linear-eq}) is the basic structure of any neural network model \cite{ref1-01}. Each linear model with a non-linear activation function is called a neuron, and usually, every layer of FNN or ANN in general, consists of a few or maybe hundreds of neurons.  Each neuron in a layer is connected to every neuron in the previous and next layers (Fig. ~\ref{fig3-1}). These are called dense or fully connected layers. FNN usually consist of input layer, a few hidden layer, and an output layer (Fig. ~\ref{fig3-1}). The information flows and processes in one direction, from the input layer through the hidden layers to the output layer, without any loops or cycles. 

\begin{align}
\label{linear-eq}
f(\textbf{x}, \textbf{w}) &= x_1 \cdot w_1 + x_2 \cdot w_2 + ... + x_n \cdot w_n + b = \textbf{x}^{T} \textbf{w} + b
\end{align}

\subsection{Multivariate Long-Short Term Memory (LSTM)}

\begin{figure}[ht]
\centering
\includegraphics[width=0.7\linewidth]{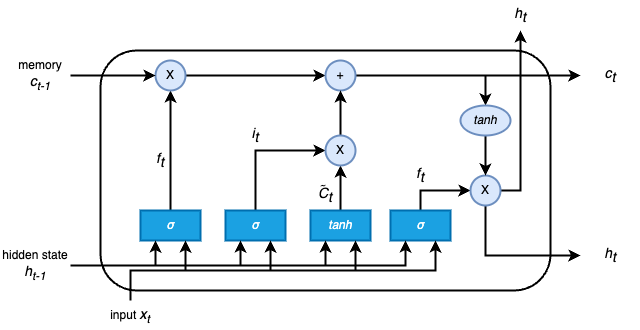}
\caption{LSTM mechanism.}
\label{fig3-2}
\vspace{-36pt} 
\end{figure}

Long Short-Term Memory (LSTM) is a Recurrent Neural Network (RNN) that differs from traditional RNN. LSTM cell uses a state that represents a "memory" or a "context" besides the inputs and the outputs \cite{ref3-01}, which aims to learn when to remember and when to forget pertinent information. LSTM contains three gates to control the dependencies, as shown in Fig.~\ref{fig3-2} an input gate to select the inputs, a forget gate to free some part of the memory, and an output gate to control the output. LSTM solve the vanishing gradient problem of traditional RNN when dealing with long sequences of data due to the application of Back Propagation Through Time (BPTT) for a specific Horizon \cite{ref3-02}. LSTM consist of these operations.

\begin{align}
\label{lstm-eq}
f_t &= \sigma(W_f \cdot [h_{t-1}, x_t] + b_f) \\
i_t &= \sigma(W_i \cdot [h_{t-1}, x_t] * b_i) \\
\tilde{C}_t &= \tanh(W \cdot [h_{t-1}, x_t] + b_C) \\
C_t &= f_t * C_{t-1} + i_t * \tilde{C}_t \\
o_t &= \sigma(W_o \cdot [h_{t-1}, x_t] * b_o) \\
h_t &= o_t * \tanh(C_t)
\end{align}

\subsection{Encoder-Decoder Architecture}
In the Encoder-Decoder architecture, there are two different inputs, the past feature values and the known current feature values. The Encoder part compresses the information from the first input sequence into a vector, which is generated from the sequence of the LSTM hidden states \cite{ref3-03}. The encoder hidden states and also the second input feed into the decoder part and generate the output sequences. In addition, some dense layers were provided before the output layer to give a better prediction sequence. Fig.  provide better explanation for the encoder-decoder architecture. This architecture also help learn the time-dependent characteristics of the sequence, give better prediction for future value \cite{ref3-04}.

\begin{figure}[ht]
\centering
\includegraphics[width=0.7\linewidth]{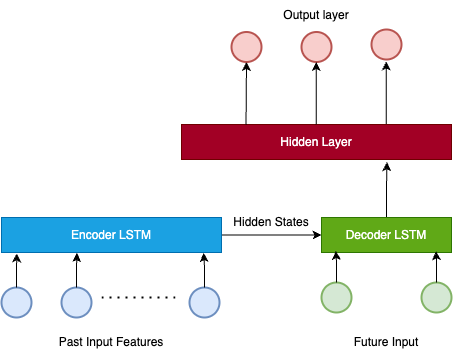}
\caption{Encoder-Decoder Architecture.}
\label{fig3-3}
\vspace{-18pt} 
\end{figure}

\subsection{Evaluation Metrics}
The models' performance in the validation dataset was assessed to determine the optimal model architecture for our research objective. The metrics used in this study as follows.
\subsubsection{Mean Squared Error (MSE)}

Mean Squared Error, abbreviated as MSE, is a fundamental metric in regression analysis that calculates the average of the squared differences between predicted values and actual observations. It provides insight into the precision of a predictive model by quantifying the average squared error across all data points. Smaller MSE values are desirable as they indicate a model that predicts closer to the actual values.
\begin{align}
MSE = \frac{1}{n} \sum_{i=1}^n (y_i - \hat{y}_i)^2 
\end{align}
    
\subsubsection{Mean Absolute Error (MAE)}

Mean Absolute Error, abbreviated as MAE, is another widely used metric in regression analysis that measures the average of the absolute differences between predicted values and actual observations. MAE provides a straightforward way to understand the average magnitude of errors made by a predictive model. Unlike Mean Squared Error (MSE), MAE does not penalize large errors as heavily, making it more robust to outliers.
\begin{align}
MAE = \frac{1}{n} \sum_{i=1}^n |y_i - \hat{y}_i| 
\end{align}

\subsubsection{Nash-Sutcliffe Efficiency (NSE)}
Nash-Sutcliffe Efficiency, assesses the goodness of fit between observed and simulated data. With values ranging from negative to 1, NSE quantifies the performance of a model in replicating observed data. An NSE of 1 represents a perfect match between model predictions and observed data, while values greater than 0 indicate that the model is superior to using the mean of the observed data. Conversely, negative NSE values suggest that the model performs worse than a basic mean-based estimate.
\begin{align}
NSE = 1 - \frac{\sum_{i=1}^n (y_i - \hat{y}_i)^2}{\sum_{i=1}^n (y_i - \bar{y})^2} 
\end{align}

\section{Dataset}
\subsection{Dataset Overview}
\begin{figure}[ht]
\centering
\includegraphics[width=0.99\linewidth]{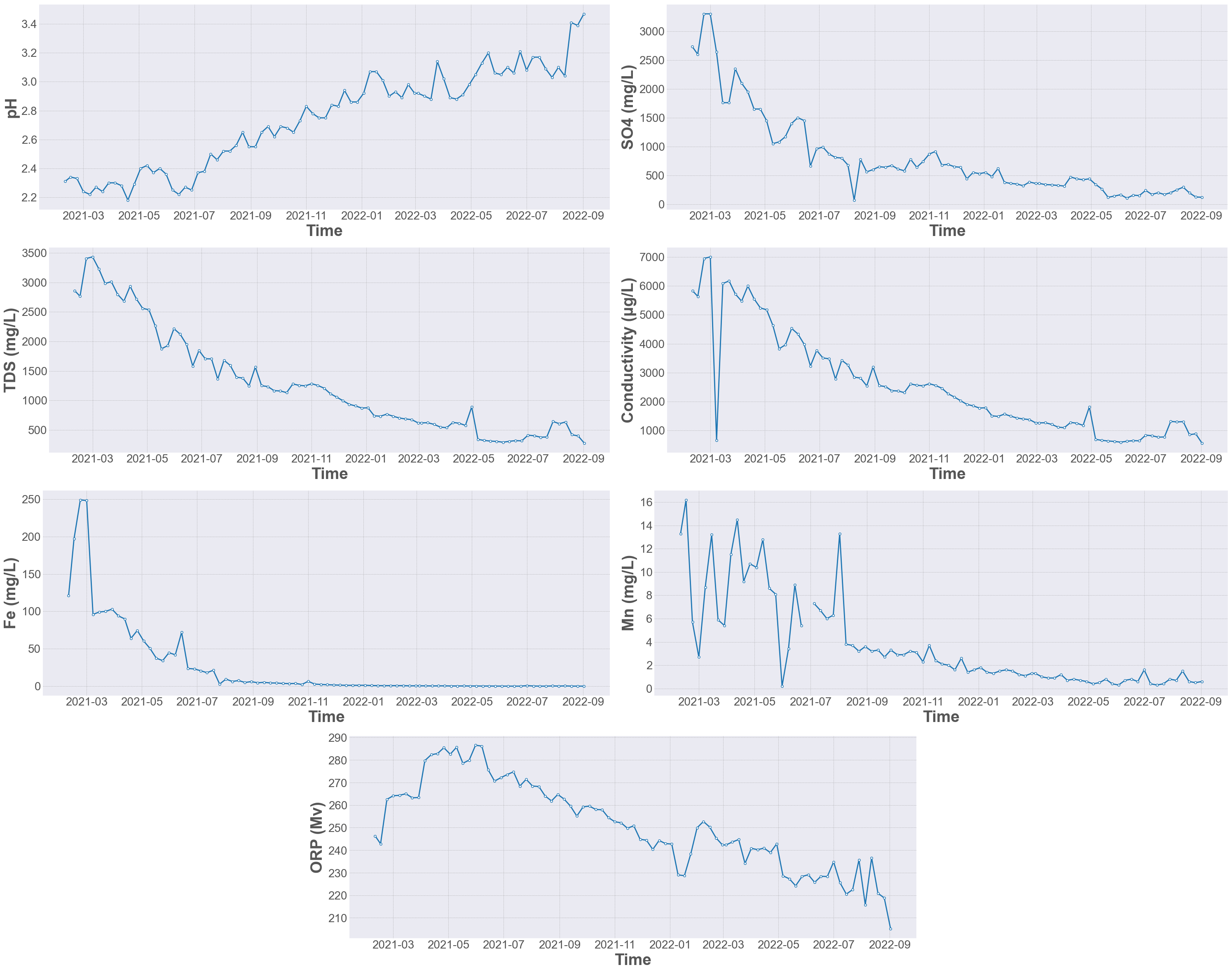}
\caption{Dataset visualization.}
\label{fig4-1}
\vspace{-18pt} 
\end{figure}

\vspace{-2pt}
The dataset for this study was acquired from one of the mining locations in Indonesia. Data was gathered between 09 February 2021 and 02 September 2022 every 7 days. This data contains 7 parameters, i.e. pH, redox potential (ORP), conductivity, total dissolved solids (TDS), SO$_4$, Fe, and Mn. Fig.~\ref{fig4-1} shows the graph visualizations of the collected data. 

\subsection{Stationarity Test}
\vspace{-3pt}
Stationarity test is an important process in time series analysis and forecasting. A stationary condition of a time series data is when its properties do not depend on the time at which the data is observed. Stationary data does not have trends and seasonality, thus making it easier to analyze and forecast \cite{ref4-01}. Therefore, the Augmented Dickey-Fuller (ADF) test was used to test the stationarity of the dataset. The test was developed using a highly significant \textit{p}-value (0.05) (Table ~\ref{tab4-01}). Additionally, line plots for the dataset were drawn to visualize and help identify any stationarity or non-stationarity properties of the dataset. 

\begin{table}
\centering
\normalfont
\caption{ADF test for stationarity.}
\label{tab4-01}       
\begin{tabular}{lll}
\hline
\multicolumn{1}{c}{\textbf{Parameter}} & \multicolumn{1}{c}{\textbf{ADF statistic}} & \multicolumn{1}{c}{\textbf{\textit{p}-value}} \\ \hline
\multicolumn{1}{c}{pH} & \multicolumn{1}{c}{-0.1143} & \multicolumn{1}{c}{0.9479} \\
\multicolumn{1}{c}{ORP} & \multicolumn{1}{c}{0.2394} & \multicolumn{1}{c}{0.9743} \\
\multicolumn{1}{c}{Conductivity} & \multicolumn{1}{c}{-4.3152} & \multicolumn{1}{c}{0.0004} \\
\multicolumn{1}{c}{TDS} & \multicolumn{1}{c}{3.9183} & \multicolumn{1}{c}{0.0019} \\
\multicolumn{1}{c}{SO$_4$} & \multicolumn{1}{c}{-4.0869} & \multicolumn{1}{c}{0.0010} \\
\multicolumn{1}{c}{Fe} & \multicolumn{1}{c}{-6.0566} & \multicolumn{1}{c}{0.0000} \\
\multicolumn{1}{c}{Mn} & \multicolumn{1}{c}{-3.9184} & \multicolumn{1}{c}{0.0019} \\ \hline

\end{tabular}
\end{table}


According to the results, pH and ORP show non-stationary behavior as indicated by their p-values exceeding the critical threshold of 0.05. On the other hand, conductivity, TDS, SO$_4$, Fe, and Mn indicates stationarity, supported by their p-values below 0.05. This means that these parameters showcased consistent properties over time, devoid of trends or seasonality, making them more amenable to analysis and forecasting. Thus, the ADF test result indicates that the dataset is stationary.

\subsection{Anomaly Detection}
\begin{figure}[ht]
\centering
\includegraphics[width=0.99\linewidth]{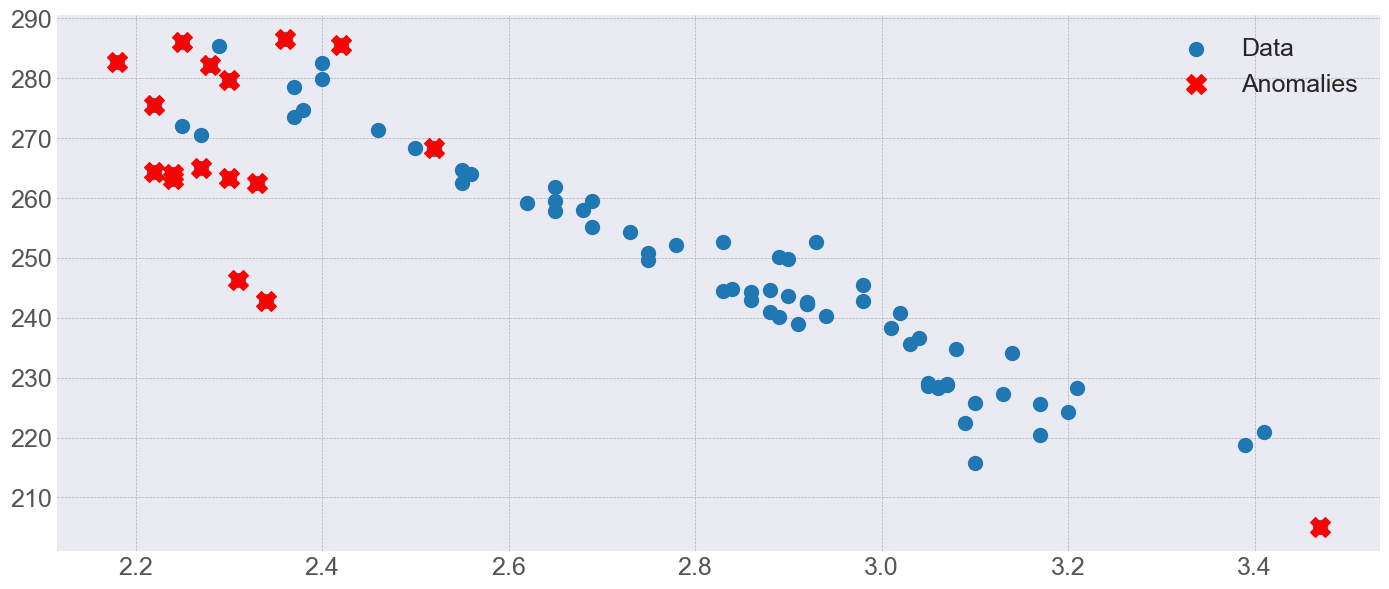}
\caption{Anomaly detection using Isolation Forest. Feature 1 (x-axis), Feature 2 (y-axis)}
\label{fig4-2}
\vspace{-12pt} 
\end{figure}

\vspace{-3pt}
Anomalies are patterns in data that have different characteristics from expected conditions and often give bias to the data \cite{ref4-02}. These anomalies can occur due to several factors, e.g., contamination, device calibration, human error, or other factors. Detecting anomalies has significant relevance and often improves performance during analysis and forecasting. Isolation forest, a model-based approach, is used for detecting anomalies, which constructs an ensemble of tree structures where anomalies are closer to the root, while normal points are deeper. This approach effectively detects anomalies with a small number of trees and minimal sub-sampling size, thus quick to convergence\cite{ref4-03}. The isolation forest was implemented with contamination parameter of 0.2 (Fig.~\ref{fig4-2}).

As indicated by their indices (0, 1, 2, 3, 4, 6, 8, 9, 10, 11, 12, 13, 16, 17, 18, 25, 82), anomalies were successfully identified using the isolation forest model with chosen contamination parameter of 0.2. These anomalies are characterized by different patterns characteristic from the expected conditions due to several factors. This successful anomaly detection is pivotal for improving the reliability and accuracy of our analysis and forecasting processes, ensuring a more robust and precise interpretation of the underlying patterns in the data.

\subsection{Data Interpolation and Transformation}
\subsubsection{Data Interpolation}
Huge amounts of data are usually needed when training ANN or deep learning \cite{ref4-04}. However, this dataset only contains 83 data points and gathered between 09 February 2021 and 02 September 2022 every 7 days. Predictive interpolation using random forest and tree-based gradient-boosting regression models were used to interpolate the missing data between the time intervals. Every model for each parameter were developed based on the time component with sine and cosine transformation. The data were split into training (80\%) and testing (20\%) sets. Each parameter is calculated based on the average of the top-3 interpolation results from the best-performing model (Table ~\ref{tab4-02}). Except for Mn parameter, where the value calculated from the result of random forest, XGBoost, and ExtraTrees.

\begin{table}
\centering
\normalfont
\caption{Predictive interpolation model performances.}
\label{tab4-02}       
\begin{tabular}{l|llllllll}
\hline
\multicolumn{1}{c|}{} & \multicolumn{8}{c}{\textbf{Model type}} \\ \cline{2-9}
\multicolumn{1}{c|}{\textbf{Parameter}} & \multicolumn{2}{c}{Random Forest} & \multicolumn{2}{c}{XGBoost} & \multicolumn{2}{c}{LigthGBM} & \multicolumn{2}{c}{ExtraTrees} \\
\multicolumn{1}{c|}{\textbf{}} & \multicolumn{1}{c}{MSE} & \multicolumn{1}{c}{MAE} & \multicolumn{1}{c}{MSE} & \multicolumn{1}{c}{MAE} & \multicolumn{1}{c}{MSE} & \multicolumn{1}{c}{MAE} & \multicolumn{1}{c}{MSE} & \multicolumn{1}{c}{MAE} \\ \hline

\multicolumn{1}{c|}{pH} & \multicolumn{1}{c}{0.0006} & \multicolumn{1}{c}{0.0244} & \multicolumn{1}{c}{0.0008} & \multicolumn{1}{c}{0.0286} & \multicolumn{1}{c}{0.0015} & \multicolumn{1}{c}{0.0393} & \multicolumn{1}{c}{0.0008} & \multicolumn{1}{c}{0.0285} \\ 

\multicolumn{1}{c|}{ORP} & \multicolumn{1}{c}{0.0004} & \multicolumn{1}{c}{0.0204} & \multicolumn{1}{c}{0.0005} & \multicolumn{1}{c}{0.0224} & \multicolumn{1}{c}{0.0010} & \multicolumn{1}{c}{0.0311} & \multicolumn{1}{c}{0.0004} & \multicolumn{1}{c}{0.0201} \\ 

\multicolumn{1}{c|}{Conductivity} & \multicolumn{1}{c}{0.0041} & \multicolumn{1}{c}{0.0644} & \multicolumn{1}{c}{0.0025} & \multicolumn{1}{c}{0.0500} & \multicolumn{1}{c}{0.0109} & \multicolumn{1}{c}{0.1044} & \multicolumn{1}{c}{0.0069} & \multicolumn{1}{c}{0.0832} \\ 

\multicolumn{1}{c|}{TDS} & \multicolumn{1}{c}{0.0042} & \multicolumn{1}{c}{0.0647} & \multicolumn{1}{c}{0.0025} & \multicolumn{1}{c}{0.0504} & \multicolumn{1}{c}{0.0112} & \multicolumn{1}{c}{0.1060} & \multicolumn{1}{c}{0.0070} & \multicolumn{1}{c}{0.0834} \\ 

\multicolumn{1}{c|}{SO$_4$} & \multicolumn{1}{c}{0.0073} & \multicolumn{1}{c}{0.0854} & \multicolumn{1}{c}{0.0057} & \multicolumn{1}{c}{0.0755} & \multicolumn{1}{c}{0.0130} & \multicolumn{1}{c}{0.1138} & \multicolumn{1}{c}{0.0107} & \multicolumn{1}{c}{0.1034} \\ 

\multicolumn{1}{c|}{Fe} & \multicolumn{1}{c}{0.0015} & \multicolumn{1}{c}{0.0381} & \multicolumn{1}{c}{0.0014} & \multicolumn{1}{c}{0.0376} & \multicolumn{1}{c}{0.0030} & \multicolumn{1}{c}{0.0545} & \multicolumn{1}{c}{0.0018} & \multicolumn{1}{c}{0.0424} \\ 

\multicolumn{1}{c|}{Mn} & \multicolumn{1}{c}{0.0127} & \multicolumn{1}{c}{0.1129} & \multicolumn{1}{c}{0.0117} & \multicolumn{1}{c}{0.1084} & \multicolumn{1}{c}{0.0084} & \multicolumn{1}{c}{0.0916} & \multicolumn{1}{c}{0.0158} & \multicolumn{1}{c}{0.1257} \\ \hline
\end{tabular}
\vspace{-12pt} 
\end{table}

\subsubsection{Data Transformation}
ANN or deep learning model usually prefer to receive inputs on the same scale. That is because ANN is just stacks of linear transforms with non linear activation function \cite{ref1-01}. Thus, building forecasting models with untransformed data often results in inaccurate forecasting results. Therefore, the data need to be transformed to close to normal distribution. The time component is also transformed using cosine transformation to give the information about time to the model.

\section{Model Development and Evaluation}
In this study, three types of model architecture, FNN, LSTM, and encoder-decoder LSTM were develop. Each of the types were trained with three different window size or the number of past time-steps data the model needs to use to predict the current time-step, except for FNN where also trained without past data. Each of the types and window sizes were trained and tested with each independent set of data. The data were split into training (70\%) and testing (30\%) sets. All of the model are also trained to forecast all of the parameters. The models were evaluate based on MSE, MAE, and NSE values. The models were trained on the same batch size of 4, ReLU activation function for the output layer, MAE loss, and adaptive moment estimation (Adam) optimizer. The result were shown on Table ~\ref{tab5-01}. All model's hyper-parameters were tuned and optimized, and callbacks were also used to create the best performing model. 

\begin{table}
\centering
\normalfont
\caption{ANN model structure and performances.}
\label{tab5-01}       
\begin{tabular}{l|llllll}
\hline
\multicolumn{1}{c|}{\textbf{Model type}} & \multicolumn{2}{c}{\textbf{Structure variation}} & \multicolumn{1}{c}{} & \multicolumn{3}{c}{\textbf{Performance metrics}} \\ 

\multicolumn{1}{c|}{\textbf{}} & \multicolumn{1}{c}{Window size} & \multicolumn{1}{c}{Epochs} & \multicolumn{1}{c}{} & \multicolumn{1}{c}{MSE} & \multicolumn{1}{c}{MAE} & \multicolumn{1}{c}{NSE} \\ \hline

\multicolumn{1}{c|}{\textbf{FNN}} & \multicolumn{1}{c}{-} & \multicolumn{1}{c}{120} & \multicolumn{1}{c}{Train} & \multicolumn{1}{c}{0.0005} & \multicolumn{1}{c}{0.0193} & \multicolumn{1}{c}{0.9952} \\

\multicolumn{1}{c|}{\textbf{}} & \multicolumn{2}{c}{} & \multicolumn{1}{c}{Validation} & \multicolumn{1}{c}{0.0011} & \multicolumn{1}{c}{0.0289} & \multicolumn{1}{c}{0.9889} \\ 

\multicolumn{1}{c|}{\textbf{}} & \multicolumn{1}{c}{7} & \multicolumn{1}{c}{80} & \multicolumn{1}{c}{Train} & \multicolumn{1}{c}{0.0006} & \multicolumn{1}{c}{0.0220} & \multicolumn{1}{c}{0.9945} \\

\multicolumn{1}{c|}{\textbf{}} & \multicolumn{2}{c}{} & \multicolumn{1}{c}{Validation} & \multicolumn{1}{c}{0.0014} & \multicolumn{1}{c}{0.0337} & \multicolumn{1}{c}{0.9861} \\ 

\multicolumn{1}{c|}{\textbf{}} & \multicolumn{1}{c}{14} & \multicolumn{1}{c}{90} & \multicolumn{1}{c}{Train} & \multicolumn{1}{c}{0.0006} & \multicolumn{1}{c}{0.0212} & \multicolumn{1}{c}{0.9945} \\

\multicolumn{1}{c|}{\textbf{}} & \multicolumn{2}{c}{} & \multicolumn{1}{c}{Validation} & \multicolumn{1}{c}{0.0011} & \multicolumn{1}{c}{0.0310} & \multicolumn{1}{c}{0.9892} \\ 

\multicolumn{1}{c|}{\textbf{}} & \multicolumn{1}{c}{28} & \multicolumn{1}{c}{120} & \multicolumn{1}{c}{Train} & \multicolumn{1}{c}{0.0004} & \multicolumn{1}{c}{0.0180} & \multicolumn{1}{c}{0.9961} \\

\multicolumn{1}{c|}{\textbf{}} & \multicolumn{2}{c}{} & \multicolumn{1}{c}{Validation} & \multicolumn{1}{c}{0.0017} & \multicolumn{1}{c}{0.0369} & \multicolumn{1}{c}{0.9837} \\ \hline

\multicolumn{1}{c|}{\textbf{LSTM}} & \multicolumn{1}{c}{7} & \multicolumn{1}{c}{125} & \multicolumn{1}{c}{Train} & \multicolumn{1}{c}{0.0005} & \multicolumn{1}{c}{0.0188} & \multicolumn{1}{c}{0.9955} \\

\multicolumn{1}{c|}{\textbf{}} & \multicolumn{2}{c}{} & \multicolumn{1}{c}{Validation} & \multicolumn{1}{c}{0.0012} & \multicolumn{1}{c}{0.0296} & \multicolumn{1}{c}{0.9883} \\ 

\multicolumn{1}{c|}{\textbf{}} & \multicolumn{1}{c}{14} & \multicolumn{1}{c}{210} & \multicolumn{1}{c}{Train} & \multicolumn{1}{c}{0.0006} & \multicolumn{1}{c}{0.0212} & \multicolumn{1}{c}{0.9943} \\

\multicolumn{1}{c|}{\textbf{}} & \multicolumn{2}{c}{} & \multicolumn{1}{c}{Validation} & \multicolumn{1}{c}{0.0010} & \multicolumn{1}{c}{0.0291} & \multicolumn{1}{c}{0.9903} \\ 

\multicolumn{1}{c|}{\textbf{}} & \multicolumn{1}{c}{28} & \multicolumn{1}{c}{225} & \multicolumn{1}{c}{Train} & \multicolumn{1}{c}{0.0004} & \multicolumn{1}{c}{0.0163} & \multicolumn{1}{c}{0.9959} \\

\multicolumn{1}{c|}{\textbf{}} & \multicolumn{2}{c}{} & \multicolumn{1}{c}{Validation} & \multicolumn{1}{c}{0.0013} & \multicolumn{1}{c}{0.0305} & \multicolumn{1}{c}{0.9871} \\ \hline

\multicolumn{1}{c|}{\textbf{Encoder-}} & \multicolumn{1}{c}{7} & \multicolumn{1}{c}{200} & \multicolumn{1}{c}{Train} & \multicolumn{1}{c}{0.0003} & \multicolumn{1}{c}{0.0150} & \multicolumn{1}{c}{0.9969} \\

\multicolumn{1}{c|}{\textbf{Decoder}} & \multicolumn{2}{c}{} & \multicolumn{1}{c}{Validation} & \multicolumn{1}{c}{0.0005} & \multicolumn{1}{c}{0.0213} & \multicolumn{1}{c}{0.9953} \\ 

\multicolumn{1}{c|}{\textbf{LSTM}} & \multicolumn{1}{c}{14} & \multicolumn{1}{c}{250} & \multicolumn{1}{c}{Train} & \multicolumn{1}{c}{0.0004} & \multicolumn{1}{c}{0.0168} & \multicolumn{1}{c}{0.9961} \\

\multicolumn{1}{c|}{\textbf{}} & \multicolumn{2}{c}{} & \multicolumn{1}{c}{Validation} & \multicolumn{1}{c}{0.0005} & \multicolumn{1}{c}{0.0206} & \multicolumn{1}{c}{0.9955} \\ 

\multicolumn{1}{c|}{\textbf{}} & \multicolumn{1}{c}{28} & \multicolumn{1}{c}{250} & \multicolumn{1}{c}{Train} & \multicolumn{1}{c}{0.0003} & \multicolumn{1}{c}{0.0164} & \multicolumn{1}{c}{0.9966} \\

\multicolumn{1}{c|}{\textbf{}} & \multicolumn{2}{c}{} & \multicolumn{1}{c}{Validation} & \multicolumn{1}{c}{0.0005} & \multicolumn{1}{c}{0.0206} & \multicolumn{1}{c}{0.9956} \\ \hline

\end{tabular}
\vspace{-12pt} 
\end{table}

\begin{figure}[ht]
\centering
\includegraphics[width=0.99\linewidth]{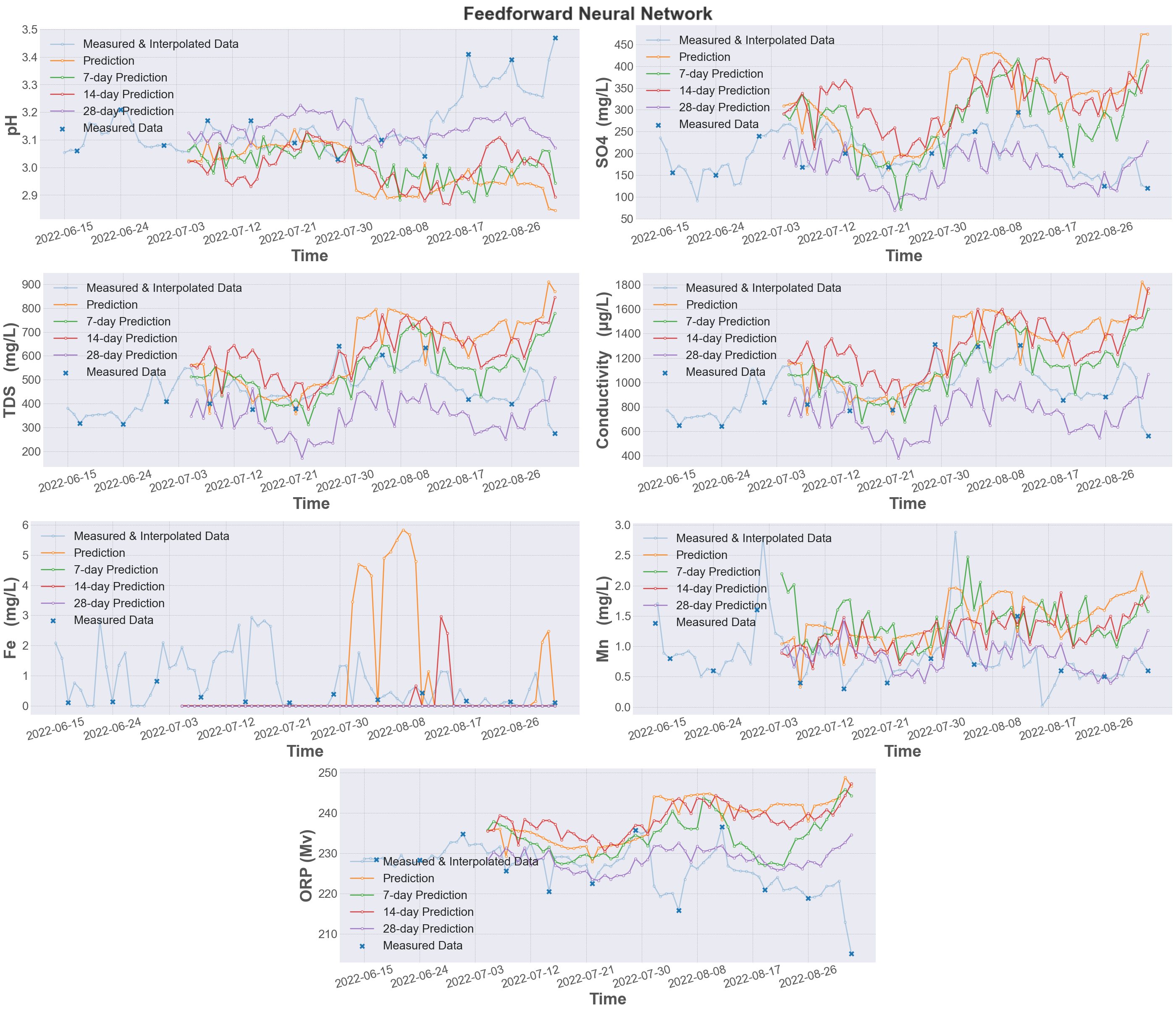}
\caption{Forecasted concentrations of all parameters for 60 days of Feedforward Neural Network. Historical data was used between 15 June 2022 to 02 September 2022 for better visualization.}
\label{fig5-01}
\vspace{-18pt} 
\end{figure}

\begin{figure}[ht]
\centering
\includegraphics[width=0.99\linewidth]{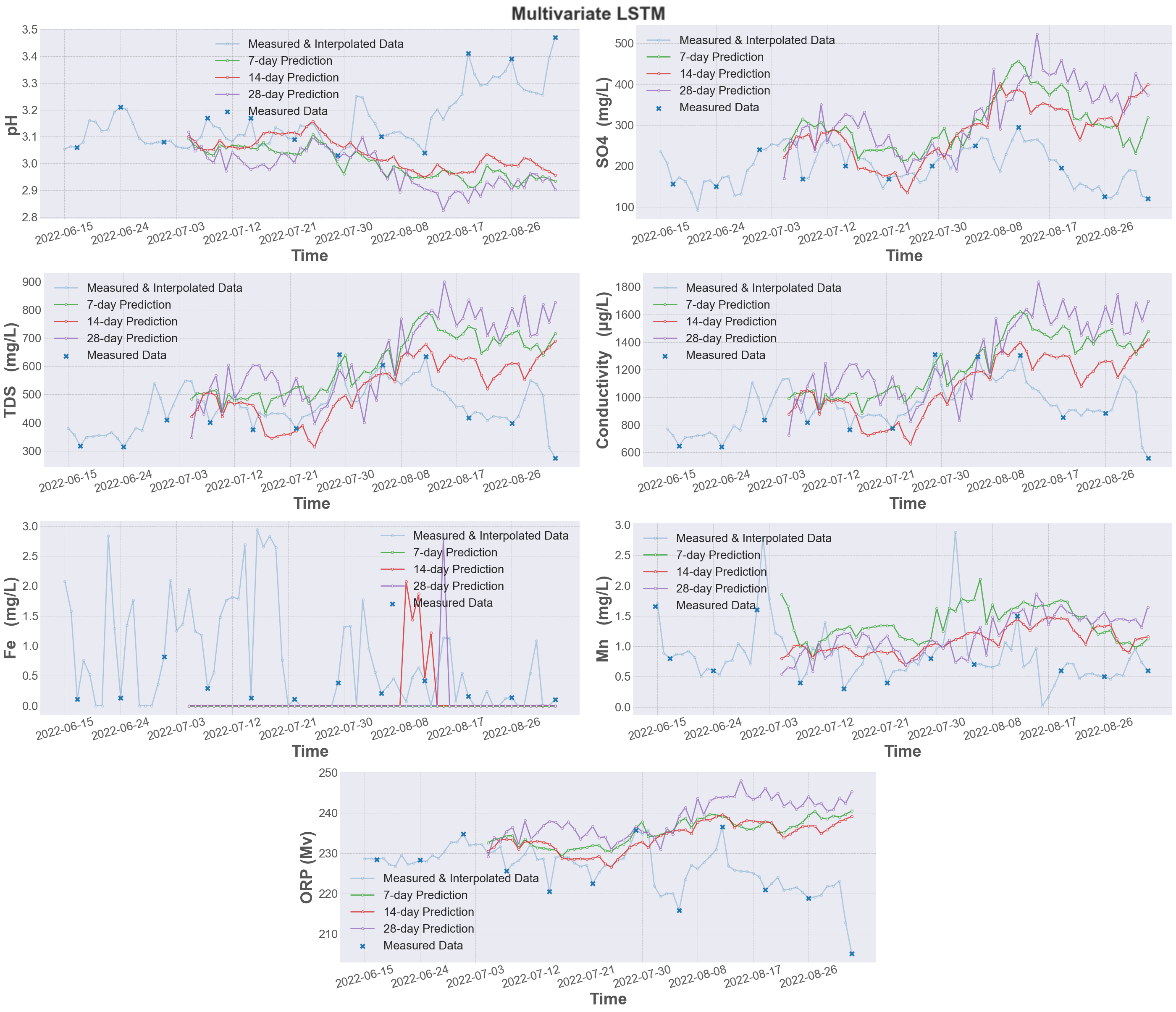}
\caption{Forecasted concentrations of all parameters for 60 days of Multivariate LSTM. Historical data was used between 15 June 2022 to 02 September 2022 for better visualization.}
\label{fig5-02}
\vspace{-18pt} 
\end{figure}

\begin{figure}[ht]
\centering
\includegraphics[width=0.99\linewidth]{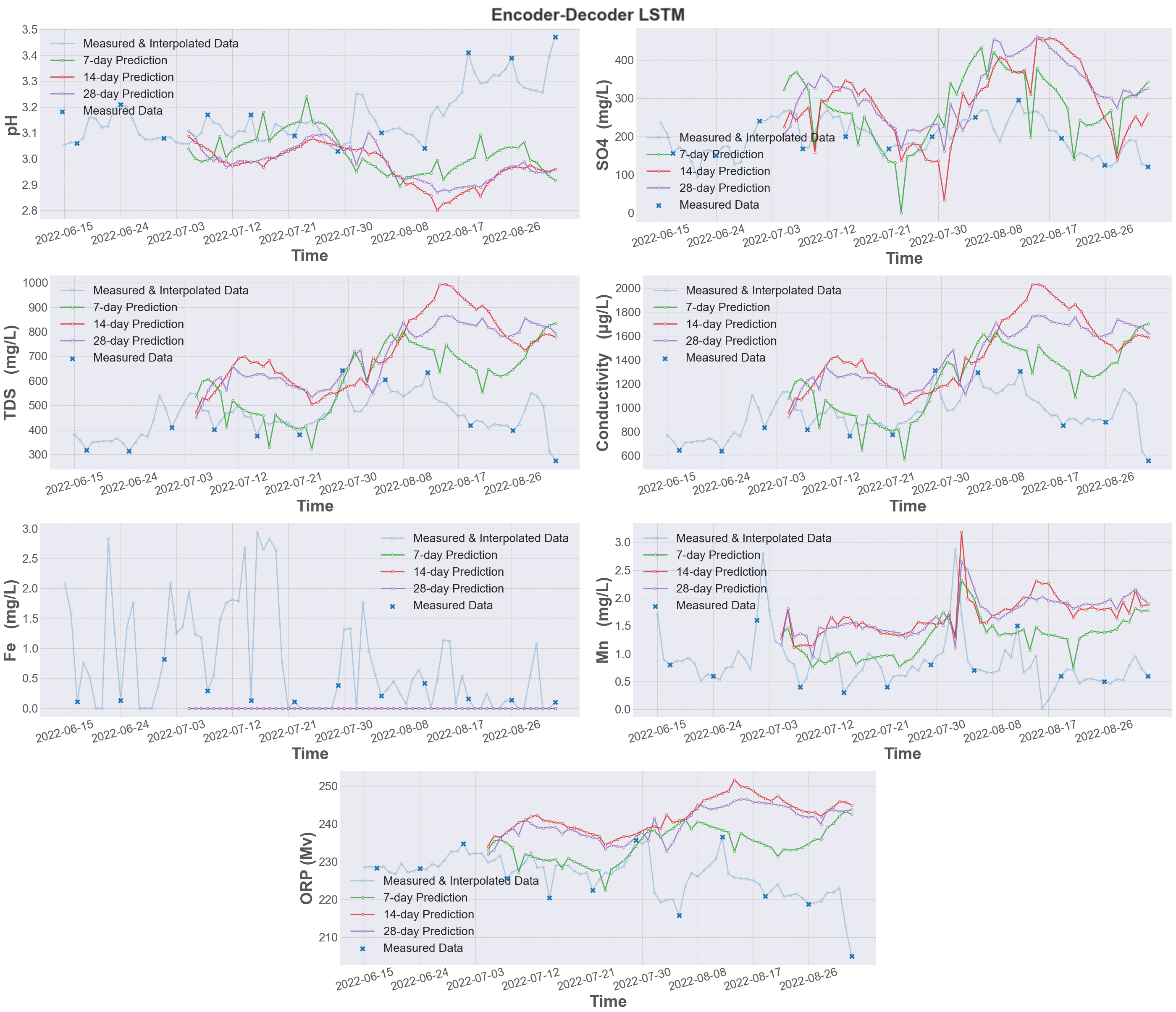}
\caption{Forecasted concentrations of all parameters for 60 days of Encoder-Decoder LSTM. Historical data was used between 15 June 2022 to 02 September 2022 for better visualization.}
\label{fig5-03}
\vspace{-18pt} 
\end{figure}

Each model was also tested to forecast the future parameter values for the next 60 days. To determine which which model is the best model of the others, the forecast result and plot given in Fig. ~\ref{fig5-01}, Fig. ~\ref{fig5-02}, and Fig. ~\ref{fig5-03} can be used, also by given the known condition of train and validation loss and also by comparing to measured data. Generally, lower training and validation loss implies good model fit on the train data and new unknown validation data. Aside from lower training and validation loss, the good and fit model is defined by the distance between training loss and the validation loss. If the distance is close enough, the model is probably good fitted, if the distance is quite far, there a possibility of overfitted model, which is not a good model, and if the validation loss is lower than the training loss, that implies the possibility of underfitted model, which also not a good model. 


\section{Result and Discussion}
From the training and validation performance, encoder-decoder LSTM displayed the overall best performance compared to other types of models and structure variation. It shows a good model fit without overfitting to the dataset, Furthermore, the variation in window size shows some overall improvement, meaning the longer the sequence of past values, give better information to the model when predicting the current value. In addition to the model's performances, the results suggest that ANN models can be applied in AMD time series forecasting and analysis. 

The models were also evaluated by forecasting all features for 60 days (Fig. ~\ref{fig5-01}, Fig. ~\ref{fig5-02}, and Fig. ~\ref{fig5-03}). Because of the kinetic test measurement and observation were not carried out daily, the measured data only contains 9 observations, while the forecasting period was for 60 days, the calculated performance and error were based on the availability of measured data. Also, based on the performance of the encoder-decoder LSTM on training and validation, we only calculate the performance of that models, for all window size (Table ~\ref{tab6-01}). Based on that result, the encoder-decoder model with 7 days of past value show better performance. This was shown by the lower MSE and MAE values compared to other window size variation. 

\begin{table}
\centering
\normalfont
\caption{MSE and MAE for forecasted data.}
\label{tab6-01}       
\begin{tabular}{lllllll}
\hline
\multicolumn{1}{c|}{\textbf{}} & \multicolumn{6}{c}{\textbf{Window Size}} \\ \cline{2-7}
\multicolumn{1}{c|}{\textbf{Parameter}} & \multicolumn{2}{c|}{7} & \multicolumn{2}{c|}{14} & \multicolumn{2}{c}{28} \\ 
\multicolumn{1}{c|}{\textbf{}} & \multicolumn{1}{c}{MSE} & \multicolumn{1}{c|}{MAE} & \multicolumn{1}{c}{MSE} & \multicolumn{1}{c|}{MAE} & \multicolumn{1}{c}{MSE} & \multicolumn{1}{c}{MAE} \\ \hline

\multicolumn{1}{c|}{pH} & \multicolumn{1}{c}{0.07} & \multicolumn{1}{c|}{0.27} & \multicolumn{1}{c}{0.09} & \multicolumn{1}{c|}{0.30} & \multicolumn{1}{c}{0.09} & \multicolumn{1}{c}{0.29}  \\

\multicolumn{1}{c|}{ORP} & \multicolumn{1}{c}{305.57} & \multicolumn{1}{c|}{17.48} & \multicolumn{1}{c}{488.37} & \multicolumn{1}{c|}{22.10} & \multicolumn{1}{c}{423.44} & \multicolumn{1}{c}{20.58} \\

\multicolumn{1}{c|}{Conductivity} & \multicolumn{1}{c}{223985.57} & \multicolumn{1}{c|}{473.27} & \multicolumn{1}{c}{387940.83} & \multicolumn{1}{c|}{622.85} & \multicolumn{1}{c}{342765.42} & \multicolumn{1}{c}{585.46} \\

\multicolumn{1}{c|}{TDS} & \multicolumn{1}{c}{56781.07} & \multicolumn{1}{c|}{238.29} & \multicolumn{1}{c}{96171.42} & \multicolumn{1}{c|}{310.12} & \multicolumn{1}{c}{85157.99} & \multicolumn{1}{c}{291.82} \\

\multicolumn{1}{c|}{SO$_4$} & \multicolumn{1}{c}{16202.92} & \multicolumn{1}{c|}{127.29} & \multicolumn{1}{c}{15768.66} & \multicolumn{1}{c|}{125.57} & \multicolumn{1}{c}{19956.18} & \multicolumn{1}{c}{141.27} \\

\multicolumn{1}{c|}{Fe} & \multicolumn{1}{c}{0.06} & \multicolumn{1}{c|}{0.24} & \multicolumn{1}{c}{0.06} & \multicolumn{1}{c|}{0.24} & \multicolumn{1}{c}{0.06} & \multicolumn{1}{c}{0.24} \\

\multicolumn{1}{c|}{Mn} & \multicolumn{1}{c}{0.66} & \multicolumn{1}{c|}{0.81} & \multicolumn{1}{c}{1.17} & \multicolumn{1}{c|}{1.08} & \multicolumn{1}{c}{1.31} & \multicolumn{1}{c}{1.14} \\ \hline

\end{tabular}
\vspace{-12pt} 
\end{table}

\section{Conclusion}
The forecasted data show low error measurements from MAE and MSE metrics and also follow the historical trend and pattern. Given this condition, the best performing model from the proposed methodology can be applied with certainty and confidence in forecasting AMD. This ANN approach show that the computer can learn patterns, trends, and seasonality of previous data in order to forecast the future value. 

This also can be concluded that by applying ANN models is a relevant contribution and addition to solve AMD problems. Finally, the results obtained in this study indicate that ANN technique are powerful and important mechanism to model and forecast the AMD data or nonlinear systems in general. These approaches also show a much better performance
and accurate approach compared to traditional time series analysis and statistical techniques. Finally, this study show the ability to predict the actual lab-scale kinetic test in order to predict the AMD with a shorter time, high accuracy, and cost efficiency.

\section*{Acknowledgments}

This study was supported by the Environment Department of PT Kaltim Prima Coal, Indonesia.

%
%
%

\end{document}